\documentclass{article}

\usepackage{arxiv}

\usepackage[utf8]{inputenc} 
\usepackage[T1]{fontenc}    
\usepackage{hyperref}       
\usepackage{url}            
\usepackage{booktabs}       
\usepackage{amsfonts}       
\usepackage{nicefrac}       
\usepackage{microtype}      
\usepackage{lipsum}		
\usepackage{graphicx}
\usepackage[numbers]{natbib}
\usepackage{doi}
\usepackage{caption}
\usepackage{subcaption}
\usepackage{amsmath}
\DeclareMathOperator*{\argmax}{argmax}
\DeclareMathOperator*{\argmin}{argmin}

\title{ALRt: An Active Learning Framework for Irregularly Sampled Temporal Data}

\author{ \href{https://orcid.org/0000-0002-0142-357X}{\includegraphics[scale=0.06]{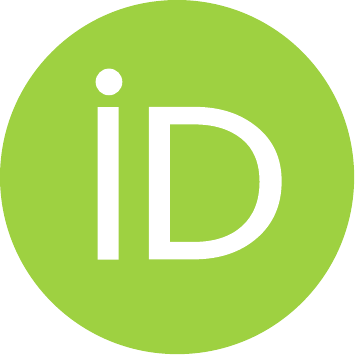}\hspace{1mm}Ronald Moore}\\
	Department of Computer Science\\
	Emory University\\
	Atlanta, GA 30329 \\
	\texttt{ronald.moore@emory.edu} \\
	\And
	\href{https://orcid.org/0000-0001-8366-4811}{\includegraphics[scale=0.06]{orcid.pdf}\hspace{1mm}Rishikesan Kamalewswaran} \\
	Department of Biomedical Informatics\\
	Emory University\\
	Atlanta, GA 30329 \\
	\texttt{rkamaleswaran@emory.edu} \\
}

\renewcommand{\shorttitle}{\textit{arXiv} Template}

\hypersetup{
pdftitle={A template for the arxiv style},
pdfsubject={q-bio.NC, q-bio.QM},
pdfauthor={David S.~Hippocampus, Elias D.~Striatum},
pdfkeywords={First keyword, Second keyword, More},
}

\begin{document}
\maketitle

\begin{abstract}
Sepsis is a deadly condition affecting many patients in the hospital. Recent studies have shown that patients diagnosed with sepsis have significant mortality and morbidity, resulting from the body's dysfunctional host response to infection. Clinicians often rely on the use of Sequential Organ Failure Assessment (SOFA), Systemic Inflammatory Response Syndrome (SIRS), and the Modified Early Warning Score (MEWS) to identify early signs of clinical deterioration requiring further work-up and treatment. However, many of these tools are manually computed and were not designed for automated computation. There have been different methods used for developing sepsis onset models, but many of these models must be trained on a sufficient number of patient observations in order to form accurate sepsis predictions. Additionally, the accurate annotation of patients with sepsis is a major ongoing challenge. In this paper, we propose the use of \underline{A}ctive \underline{L}earning \underline{R}ecurren\underline{t} Neural Networks (ALRts) for short temporal horizons to improve the prediction of irregularly sampled temporal events such as sepsis. We show that an active learning RNN model trained on limited data can form robust sepsis predictions comparable to models using the entire training dataset.
\end{abstract}


\section{Introduction}
\label{sec:introduction}
Sepsis is a deadly condition affecting many patients in the hospital. According to \cite{singer_third_2016}, sepsis is defined as a dysfunctional host response to infection causing major organ failure and increasing the risk of death or major disability. A report published by the World Health Organization (WHO) \cite{who_2020} revealed that there were roughly 49 million cases of sepsis worldwide in 2017, with approximately 20 million of these sepsis cases occurring among children less than five years old. The report stated that there were an estimated 11 million sepsis-related deaths in 2017, accounting for approximately 20\% of deaths worldwide, more than double the previous estimate.
\par Results from a study conducted from 1999 to 2005 showed that the national age-adjusted sepsis mortality rate in the U.S. was roughly 65.5 per 100,000 persons and the state-level sepsis mortality rate ranged from 41 to 88.6 per 100,000 persons \cite{wang2010}. Furthermore, the study showed that the highest sepsis mortality rates were seen in the mid-Atlantic and Southeastern regions of the U.S. In a similar study conducted from 2004-2012, the amount of pediatric severe sepsis increased in the studied U.S. children’s hospitals \cite{balamuth2014}.
\par There are widely adopted clinical criteria used to identify the risk of sepsis and prevent the onset of sepsis. Some of the most popular sepsis criteria are the Sequential Organ Failure Assessment (SOFA) \cite{jones_sequential_2009}, Systemic Inflammatory Response Syndrome (SIRS) criteria \cite{kaukonen_systemic_2015}, and the Modified Early Warning Score (MEWS) criteria \cite{yu2014comparison}. However, the substantial amount of lab vitals and long patient observation time needed for these criteria make them very expensive. Furthermore, since many of these criteria use different markers and metrics, they have conflicting predictions regarding a patient's sepsis risk. Unwin et. al \cite{unwin2021} performed a study in which they observed a subset of patients from 14 hospitals in Wales that were either in the emergency department (ED) or ward and compared sepsis prediction outcomes from the different sepsis criteria. They found that the predictions only matched 12 percent of the patients. Furthermore, when they plotted the 90-day mortality prediction receiver operating characteristic (ROC) curves for each criterion, they found that these predictions were not much better than random guesses. Indeed, there has been a large number of sepsis prediction algorithms that have been recently contributed. However, those algorithms often use surveillance definitions which were not meant for identifying patients with sepsis \cite{fleuren2020machine}. The utilization of such definitions may mean that some patients may have been unrecognized, even when they demonstrate substantial clinical similarities with septic patients. Moreover, manual labeling of sepsis by utilizing chart review can be an expensive task, requiring significant time and resources. Therefore, there is a need to utilize active learning techniques as a means to bridge the gap in identifying patients who may be similar to labeled sepsis patients in our dataset. We utilize recurrent neural networks (RNNs) to model temporal dynamics of vital signs and other clinical data and investigate whether the combination of these approaches may improve the prediction of sepsis.

\par Our main contributions can be summarized as follows:
\begin{itemize}
  \item We propose an active learning framework for irregularly sampled temporal data to improve the prediction of events with temporal short horizons such as sepsis.
  \item Experiment results show that the active learning framework performs comparably to traditional temporal frameworks with the utilization of significantly less training data.
  \item Post-experiment analysis shows the differences in feature importance rankings between the active learning framework and the traditional temporal framework.
\end{itemize}

\section{Related Work}
\label{sec:related_work}
\subsection{Sepsis Prediction}
There have been various methods that apply machine learning algorithms to the problem of sepsis prediction. The Artificial Intelligence Sepsis Expert (AISE) algorithm \cite{nemati_interpretable_2018} uses a modified regularized version of the Weilbull-Cox proportional hazard model. However, the AISE model was trained on EMR data recorded by bedside nurses, which ultimately may have introduced some bias into the model. 
\par The InSight algorithm \cite{desautels_prediction_2016} predicted the early onset of sepsis in a subset of patients from the MIMIC-III dataset. However, this algorithm uses several sequential calculations that are not possible to calculate manually. Therefore, more work has to be done to make the InSight predictions explainable to clinicians, such as providing approximate manual calculations.
\par \cite{brown_prospective_2016} used a Naive Bayes model to predict severe sepsis and septic shock in patients. However, the performance of this model is highly dependent upon the probability value selected for the alert threshold, which may have to be constantly tuned for patient distributions in different hospitals. The Sepsis Early Risk Assessment (SERA) algorithm \cite{goh_artificial_2021} leveraged both structured and unstructured data for its sepsis diagnosis and early prediction algorithms. A drawback of the SERA algorithm is that it was not tested in different hospital environments.
\par There have been some successful approaches that have used models suited for sequential data, RNNs \cite{futoma_learning_2017, futoma_improved_2017, scherpf_predicting_2019}, LSTMs \cite{zhang2021}, and TCNs \cite{moor2019, kok_automated_2020, wang_sepsis_2022}. However, these models did not incorporate any active learning techniques into their models.
\par The Online Artificial Intelligence Experts Competing Framework (OnAI-Comp) \cite{zhou2021} uses an online learning approach for sepsis prediction. The framework used a trusted third party to distribute patients' information to a group of AI experts, which can be both models and medical professionals. These experts then made sepsis predictions on each patient based on the received information. These predictions were then given to a trusted third party which decided the final predictions for each patient. A limitation of the OnAI-Comp framework is that there is a potential risk that patients might receive inaccurate sepsis prediction results during the early hours of their visit.

\subsection{Time-Series Prediction Networks}
\subsubsection{Recurrent Neural Networks}
There are various deep learning frameworks that are suitable for time-series prediction problems. One of the more popular frameworks is the recurrent neural network (RNN). They use hidden state vectors to represent the context based on prior inputs and outputs. The RNN is characterized by its ability to remember the past. As a result, its current predictions are influenced by what it has learned from past observations. RNNs can take a variable length input and produce an output vector of the same length. A disadvantage of RNNs is that they have difficulty processing very long input sequences.

\subsubsection{Long Short-Term Memory Networks}
Long short-term memory networks (LSTMs) are designed to avoid the long-term dependency problem that RNNs experience \cite{hochreiter1997}. LSTMs are special types of RNNs that are able to remember information for longer periods of time than RNNs. Similar to RNNs, LSTMs have a chain-like structure of repeating modules. However, this module consists of four different structures. These structures are the cell, the input gate, the output gate, and the forget gate. The interactions between these four structures are responsible for the long memory abilities of LSTMs. One drawback of LSTMs is that they are slower to train than traditional RNNs.

\subsubsection{Temporal Convolutional Networks}
Temporal convolutional networks (TCNs) are another alternative solution to RNNs for time-series prediction tasks \cite{lea2016}. TCNs consist of an encoder-decoder framework of spatial CNNs and pooling, upsampling, and normalization layers. A disadvantage of TCNs is that they may not be as memory efficient as RNNs since they require the raw sequence up to the effective history length as input.

\subsection{Active Learning}
Active learning is a subset of machine learning which can be thought of as a semi-supervised learning process. Whereas supervised machine learning algorithms only deal with labeled data, active learning algorithms used both labeled and unlabeled to train the model. Initially, the model is trained on a subset of the training data that is labeled. Next, the trained model is asked to give predictions on the unlabeled training data. From these predictions, the instances that the model is least certain about are then labeled by an oracle and placed into the set of labeled training data and the model resumes training. This process is repeated until all unlabeled training data has been included in the training pool or some stopping criterion is reached.

\subsubsection{Stopping Criteria}
There have been various stopping criteria suggested for the active learning process. One of the first stopping criterion methods for active learning required that the confidence of the model was evaluated on a separate test set of unlabeled data after each active learning training session \cite{vlachos_stopping_2008}. However, it was noted that this stopping criterion required a consistent improvement in the classifier's confidence, and defining this consistent improvement is not a straightforward task. Some other confidence-based stopping criteria for active learning include the max-confidence, min-error, overall-uncertainty, and classification-change heuristics \cite{zhu2010}. A stability-based stopping criterion was also suggested as an alternative to confidence-based stopping criteria \cite{wang_stability-based_2014}. Many of these stopping criteria were evaluated with more traditional probabilistic models, whereas this framework uses RNNs.

\subsubsection{Deep Active Learning Methods}
The use of active learning methods with deep learning frameworks has been investigated for applications such as image classification \cite{gal2017, wang2017} tasks and natural language processing (NLP) tasks such as named entity recognition \cite{shen2018deep} and hashtag segmentation \cite{glushkova2019}. However, this work incorporates active learning techniques with deep learning models for time-series prediction applications.

\section{Methodology}
\label{sec:methodology}
\begin{figure*}[t]
    \centering
    \includegraphics[width=0.5\textwidth]{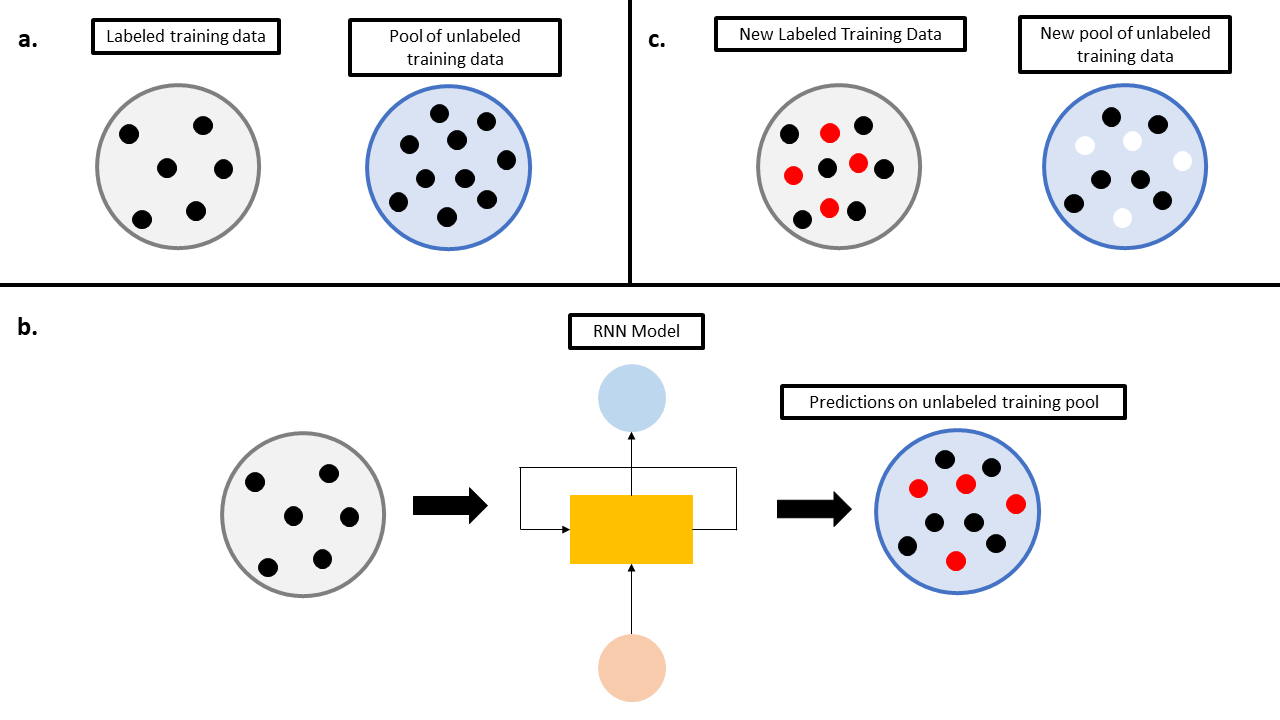}
    \caption{ALRt framework. \textbf{a.} The initial labeled training data and the pool of unlabeled training data. \textbf{b.} The labeled training data is fed into the RNN model for training. Predictions are then made on the pool of unlabeled training data. The samples which the model is the most uncertain about are colored red. \textbf{c.} The predictions with the highest uncertainty are removed from the pool of unlabeled training data and placed into the labeled training data for the next round of training.}
    \label{fig:active_rnn}
\end{figure*}

There are different methods used to determine which samples the model is least unsure about when giving its predictions. The most popular sampling methods are the least confident, margin, and entropy uncertainty sampling methods \cite{settles2010}.

\subsection{Least Confident Sampling}
Least confident sampling chooses the sample with the smallest probability value for its most probable class is selected. The logic behind this is that samples with smaller probabilities for their most likely class correspond to samples with high uncertainty. Uncertainty sampling can be represented by the equation

\begin{equation}
\label{lc_eq}
x_{lc} = \argmax_{x}\:1-p_{\Theta}(\hat{c}|x)
\end{equation}

where $p_{\Theta}(\hat{c}|x)$ represents the probability of the most probable class $\hat{c}$ for observation $x$ under the model $\Theta$.

\subsection{Margin Sampling}
Margin sampling picks the sample with the smallest difference between the two most probable classes predicted. The reasoning behind this metric is that samples that have similar probabilities among multiple classes correlate to samples with a high level of uncertainty. This can be represented by the equation

\begin{equation}
\label{marg_eq}
x_{m} = \argmin_{x}\:p_{\Theta}(\hat{c}_{1}|x) - p_{\Theta}(\hat{c}_{2}|x)
\end{equation}

where $p_{\Theta}(\hat{c}_{1}|x)$ and $p_{\Theta}(\hat{c}_{2}|x)$ represent the probabilities of the first and second most probable classes $\hat{c}_{1}$ and $\hat{c}_{2}$ for observation $x$ under the model $\Theta$ respectively.

\subsection{Entropy Sampling}
\par Entropy sampling selects the sample with the largest entropy value. Samples with higher levels of entropy correspond to samples with higher levels of uncertainty. The entropy sampling method can be represented by the equation

\begin{equation}
\label{entr_eq}
x_{e} = \argmax_{x}\:-\sum_{i}p_{\Theta}(\hat{c}_{i}|x)\:log\:p_{\Theta}(\hat{c}_{i}|x)
\end{equation}

where $p_{\Theta}(\hat{c}_{i}|x)$ represents the probability of the \textit{i}th class $\hat{c}_{i}$ for observation $x$ under the model $\Theta$. 
\par Since the sequence for each patient varies in length, then the standard uncertainty, margin, and entropy sampling methods would inherently place importance on patients with longer time sequences. To handle this problem, normalized versions of these sampling methods are used \cite{shen2018deep}. Normalized versions of equations \ref{lc_eq}, \ref{marg_eq}, and \ref{entr_eq} can now be defined by

\begin{equation}
\label{norm_lc_eq}
x_{nlc} = \argmax_{x}\frac{1}{n}\sum_{n}1-p_{\Theta}(\hat{c}|x_{n})
\end{equation}

\begin{equation}
\label{norm_marg_eq}
x_{nm} = \argmax_{x}\frac{1}{n}\sum_{n}p_{\Theta}(\hat{c}_{1}|x_{n}) - p_{\Theta}(\hat{c}_{2}|x_{n})
\end{equation}

\begin{equation}
\label{norm_entr_eq}
x_{ne} = \argmax_{x}\:-\frac{1}{n}\sum_{i}\sum_{n}p_{\Theta}(\hat{c}_{i}|x_{n})\:log\:p_{\Theta}(\hat{c}_{i}|x_{n})
\end{equation}

In equation \ref{norm_lc_eq}, $p_{\Theta}(\hat{c}|x_{n})$ represents the probability of the most probable class for observation $x$ at time step $n$ under the model $\Theta$. In equation \ref{norm_marg_eq}, $p_{\Theta}(\hat{c}_{1}|x_{n})$ and $p_{\Theta}(\hat{c}_{2}|x_{n})$ represent the probabilities of the first and second most probable classes for observation $x$ at time step $n$ under the model $\Theta$. For equation \ref{norm_entr_eq}, $p_{\Theta}(\hat{c}_{i}|x_{n})$ represents the probability of class $i$ for observation $x$ at time step $n$ under the model $\Theta$.

\section{Experiment}
\label{sec:experiment}
\subsection{Experiment Design}
For the experiment, the performances of active learning RNN models trained on limited training data were compared to an RNN model trained on the full training dataset. A total of 15 active learning models were created, each of which were trained on different fractions of the dataset. For the rest of the paper, Active RNN\_$N$ will be used to denote an ALRt model trained on $N$ percent of the training data. For example, Active RNN\_20 was trained on only 20 percent of the training data.

\subsection{Dataset Preparation}
The dataset used for the experiment came from the 2019 PhysioNet Challenge \cite{reyna2020}. The original dataset contains 37,404 nonseptic patients and 2,932 septic patients, for a total of 40,336 patients. Any patients with less than 24 hours of data were excluded from the final dataset. The final dataset contains 28,817 nonseptic patients and 1844 septic patients, for a total of 30,661 patients. In order to handle this dataset imbalance of nonseptic and septic patients, class importance weighting was used for all models in the experiment. The training and test sets were divided using a stratified approach at the patient level in order to prevent contamination between the two datasets.

\subsection{Feature Selection and Engineering}
The 2019 PhysioNet Challenge Dataset contains 8 vital signs, 26 laboratory values, and 6 demographic markers, for a total of 40 features. All vital signs and laboratory values were included in the final feature set. All demographic variables were excluded from the final feature set in order to mitigate bias. In regards to feature engineering, two features were created before dataset imputation in order to count the number of labs ordered for a patient over 12-hour and 48-hour periods. The final feature set consisted of 8 vital signs, 26 laboratory values, and 2 feature-engineered variables, for a total of 36 features. In terms of imputation, missing vital signs and laboratory values were forward-filled up to 12 hours and 36 hours respectively. Any remaining missing values were replaced with the population median for vital signs and laboratory values respectively. All features were standardized by first removing the mean and then scaling to unit variance. More information on the final feature set and other dataset statistics can be found in Table \ref{tab:table_one}.

\subsection{Model Selection and Model Training}
We utilized an RNN for the baseline model and active learning models. We chose an RNN because it is known to work well with sequences of varying lengths as well as time-series data. Similar to \cite{glushkova2019}, it was chosen to start training the model with only a subset of the training data and add more training data after each epoch using active learning sampling methods. The active learning recurrent neural network (ALRt) framework can be seen in Figure \ref{fig:active_rnn}.
\par In order to remove randomness from the experiment, a $k$-fold cross-validation was performed on the dataset. The dataset was split into 5 folds. At each cross-validation round, 1 fold was treated as the test set and the remaining 4 folds were used as the training set.
\par The ALRt model was initially given access to 20 percent of the training data, while the remaining 80 percent of training data was treated as unlabeled data. After each round, another 20 percent of data was added to the training pool, until finally all training data was used. This active learning process was repeated for the uncertainty, margin, and entropy-based sampling methods respectively. The simple RNN model was trained for 5 epochs. The number of epochs was chosen to be 5 since the active learning process took a total of 5 iterations to include the full training dataset.

\subsection{Evaluation Metrics}
In addition to measuring the accuracy of the models, the specificity, sensitivity, precision, area under the receiver operating characteristic curve (AUROC), and area under the precision-recall curve (AUPRC) were also measured for the models. These metrics were included due to the heavily imbalanced nature of the dataset.

\section{Results}
\label{sec:results}
\subsection{Sepsis and Nonsepsis Cohorts}
\begin{table*}[t]
    \centering
    \resizebox{\textwidth}{!}{
    \begin{tabular}{llllllll}
    \toprule
     &  &  \textbf{Missing} &  \textbf{Overall} & \textbf{No Sepsis} & \textbf{Sepsis} & P-Value & \textbf{Test} \\
    \midrule
    \textbf{n} & {} &  &  30661 & 28817 & 1844 &  & \\
    \textbf{Age}, median [Q1,Q3] &     &                 0 &     64.0 [52.0,74.4] &     64.0 [51.9,74.4] &     64.8 [52.1,74.1] &   0.599 &  Kruskal-Wallis \\
    \textbf{Gender}, n (\%) & Female &                 0 &         13540 (44.2) &         12801 (44.4) &           739 (40.1) &  <0.001 &     Chi-squared \\
    & Male &                   &         17121 (55.8) &         16016 (55.6) &          1105 (59.9) &         &                 \\
    \textbf{Hospital Admission Time}, median [Q1,Q3] &     &                 0 &    -5.9 [-45.2,-0.0] &    -6.0 [-44.5,-0.0] &    -3.3 [-66.5,-0.0] &  <0.001 &  Kruskal-Wallis \\
    \textbf{ICU Length of Stay}, median [Q1,Q3] &     &                 0 &     43.0 [37.0,50.0] &     42.0 [37.0,49.0] &    67.0 [42.0,113.0] &  <0.001 &  Kruskal-Wallis \\
    \textbf{MICU}, n (\%) & No &             11867 &          9449 (50.3) &          9003 (50.5) &           446 (46.0) &   0.007 &     Chi-squared \\
     & Yes &                   &          9345 (49.7) &          8821 (49.5) &           524 (54.0) &         &                 \\
    \textbf{SICU}, n (\%) & No &             11867 &          9345 (49.7) &          8821 (49.5) &           524 (54.0) &   0.007 &     Chi-squared \\
    & Yes &                   &          9449 (50.3) &          9003 (50.5) &           446 (46.0) &         &                 \\
    \textbf{Heart Rate}, median [Q1,Q3] &     &                 2 &   103.0 [91.0,116.0] &   102.0 [90.0,115.0] &  114.0 [101.0,128.0] &  <0.001 &  Kruskal-Wallis \\
    \textbf{Pulse Oximetry (\%)}, median [Q1,Q3] &     &                 4 &  100.0 [100.0,100.0] &  100.0 [100.0,100.0] &  100.0 [100.0,100.0] &  <0.001 &  Kruskal-Wallis \\
    \textbf{Temperature (Celsius)}, median [Q1,Q3] &     &               121 &     37.5 [37.1,38.0] &     37.5 [37.1,37.9] &     38.1 [37.6,38.7] &  <0.001 &  Kruskal-Wallis \\
    \textbf{Systolic Blood Pressure}, median [Q1,Q3] &     &               218 &  153.0 [137.0,170.0] &  152.0 [137.0,170.0] &  161.0 [142.5,180.0] &  <0.001 &  Kruskal-Wallis \\
    \textbf{Mean Arterial Pressure}, median [Q1,Q3] &     &                41 &   104.0 [94.0,117.4] &   104.0 [93.3,117.0] &   110.0 [97.0,125.0] &  <0.001 &  Kruskal-Wallis \\
    \textbf{Diastolic Blood Pressure}, median [Q1,Q3] &     &              5402 &     83.0 [72.5,95.0] &     82.0 [72.0,95.0] &     86.0 [75.0,98.0] &  <0.001 &  Kruskal-Wallis \\
    \textbf{Respiration Rate}, median [Q1,Q3] &     &                29 &     26.0 [23.0,30.0] &     26.0 [23.0,30.0] &     30.0 [25.0,34.9] &  <0.001 &  Kruskal-Wallis \\
    \textbf{End Tidal CO2}, median [Q1,Q3] &     &             27865 &     38.0 [33.0,43.0] &     38.0 [33.0,43.0] &     39.0 [34.0,44.0] &   0.104 &  Kruskal-Wallis \\
    \textbf{Base Excess}, median [Q1,Q3] &     &             19699 &        1.0 [0.0,3.3] &        1.0 [0.0,3.0] &        2.0 [0.0,6.0] &  <0.001 &  Kruskal-Wallis \\
    \textbf{Bicarbonate (HCO3)}, median [Q1,Q3] &     &             14569 &     26.0 [23.0,28.0] &     26.0 [23.0,28.0] &     26.0 [24.0,29.0] &  <0.001 &  Kruskal-Wallis \\
    \textbf{FiO2 (\%)}, median [Q1,Q3] &     &             15932 &        0.7 [0.5,1.0] &        0.6 [0.5,1.0] &        1.0 [0.5,1.0] &  <0.001 &  Kruskal-Wallis \\
    \textbf{pH}, median [Q1,Q3] &     &             14995 &        7.4 [7.4,7.5] &        7.4 [7.4,7.5] &        7.5 [7.4,7.5] &  <0.001 &  Kruskal-Wallis \\
    \textbf{Partial Pressure of Carbon Dioxide (PaCO2)}, median [Q1,Q3] &     &             15453 &     44.0 [39.0,50.0] &     44.0 [39.0,49.8] &     46.0 [40.0,53.0] &  <0.001 &  Kruskal-Wallis \\
    \textbf{Oxygen Saturation (\%)}, median [Q1,Q3] &     &             19803 &     98.0 [97.0,99.0] &     98.0 [97.0,99.0] &     98.0 [97.0,99.0] &  <0.001 &  Kruskal-Wallis \\
    \textbf{Aspartate Aminotransferase (AST)}, median [Q1,Q3] &     &             18767 &     35.0 [21.0,76.0] &     34.0 [21.0,72.0] &    49.0 [27.0,109.5] &  <0.001 &  Kruskal-Wallis \\
    \textbf{Blood Urea Nitrogen (BUN)}, median [Q1,Q3] &     &               806 &     19.0 [13.0,30.0] &     18.0 [13.0,29.0] &     26.0 [16.0,43.0] &  <0.001 &  Kruskal-Wallis \\
    \textbf{Alkaline Phosphatase}, median [Q1,Q3] &     &             18919 &    73.0 [55.0,106.0] &    73.0 [55.0,106.0] &    80.0 [58.0,119.0] &  <0.001 &  Kruskal-Wallis \\
    \textbf{Calcium}, median [Q1,Q3] &     &              2932 &        8.6 [8.2,9.0] &        8.6 [8.2,9.0] &        8.6 [8.2,9.1] &  <0.001 &  Kruskal-Wallis \\
    \textbf{Chloride}, median [Q1,Q3] &     &             13568 &  108.0 [104.0,111.0] &  107.0 [104.0,111.0] &  109.0 [105.0,113.0] &  <0.001 &  Kruskal-Wallis \\
    \textbf{Creatinine}, median [Q1,Q3] &     &               812 &        1.0 [0.8,1.4] &        1.0 [0.8,1.4] &        1.2 [0.8,2.0] &  <0.001 &  Kruskal-Wallis \\
    \textbf{Direct Bilirubin}, median [Q1,Q3] &     &             28843 &        0.3 [0.1,1.1] &        0.3 [0.1,1.0] &        0.6 [0.2,2.4] &  <0.001 &  Kruskal-Wallis \\
    \textbf{Glucose}, median [Q1,Q3] &     &               703 &  154.0 [126.0,194.0] &  153.0 [125.0,192.0] &  175.0 [145.0,220.0] &  <0.001 &  Kruskal-Wallis \\
    \textbf{Lactic Acid}, median [Q1,Q3] &     &             20083 &        1.9 [1.3,3.1] &        1.9 [1.3,3.1] &        2.1 [1.4,3.7] &  <0.001 &  Kruskal-Wallis \\
    \textbf{Magnesium}, median [Q1,Q3] &     &              2517 &        2.1 [2.0,2.4] &        2.1 [2.0,2.4] &        2.3 [2.1,2.5] &  <0.001 &  Kruskal-Wallis \\
    \textbf{Phosphate}, median [Q1,Q3] &     &              7511 &        3.7 [3.1,4.4] &        3.6 [3.0,4.4] &        4.0 [3.3,4.9] &  <0.001 &  Kruskal-Wallis \\
    \textbf{Potassium}, median [Q1,Q3] &     &               787 &        4.3 [4.0,4.7] &        4.3 [4.0,4.7] &        4.5 [4.2,5.0] &  <0.001 &  Kruskal-Wallis \\
    \textbf{Total Bilirubin}, median [Q1,Q3] &     &             18848 &        0.8 [0.5,1.4] &        0.8 [0.5,1.3] &        1.0 [0.6,1.9] &  <0.001 &  Kruskal-Wallis \\
    \textbf{Troponin I}, median [Q1,Q3] &     &             25250 &        0.2 [0.0,2.2] &        0.1 [0.0,2.2] &        0.3 [0.1,2.4] &   0.003 &  Kruskal-Wallis \\
    \textbf{Hematocrit (\%)}, median [Q1,Q3] &     &               981 &     33.2 [30.0,37.2] &     33.2 [30.0,37.2] &     33.4 [30.2,37.2] &   0.093 &  Kruskal-Wallis \\
    \textbf{Hemoglobin}, median [Q1,Q3] &     &              1018 &      11.1 [9.9,12.4] &      11.1 [9.9,12.4] &     11.1 [10.0,12.5] &   0.054 &  Kruskal-Wallis \\
    \textbf{Partial Thromboplastin Time (PTT)}, median [Q1,Q3] &     &             13931 &     32.7 [28.0,43.5] &     32.5 [27.9,43.0] &     34.7 [29.2,51.3] &  <0.001 &  Kruskal-Wallis \\
    \textbf{White Blood Cell Count}, median [Q1,Q3] &     &              1077 &      11.5 [8.4,15.2] &      11.3 [8.3,15.0] &     14.5 [10.9,18.8] &  <0.001 &  Kruskal-Wallis \\
    \textbf{Fibrinogen}, median [Q1,Q3] &     &             26688 &  292.0 [214.0,406.0] &  287.0 [211.0,396.0] &  339.0 [239.0,498.5] &  <0.001 &  Kruskal-Wallis \\
    \textbf{Platelets}, median [Q1,Q3] &     &              1056 &  206.0 [154.0,270.0] &  205.0 [154.0,269.0] &  219.5 [157.0,293.0] &  <0.001 &  Kruskal-Wallis \\
    \textbf{Labs Ordered in Last 12 Hours}, median [Q1,Q3] &     &                 0 &     25.0 [16.0,39.0] &     24.0 [16.0,37.0] &     38.0 [26.0,55.0] &  <0.001 &  Kruskal-Wallis \\
    \textbf{Labs Ordered in Last 48 Hours}, median [Q1,Q3] &     &                 0 &     45.0 [29.0,72.0] &     43.0 [28.0,69.0] &    79.0 [52.0,118.0] &  <0.001 &  Kruskal-Wallis \\
    \bottomrule
    \end{tabular}
    }
    \caption{Cohort statistics for the included patients from the 2019 PhysioNet Challenge dataset. For aggregation, the maximum available value for a patient was used.}
    \label{tab:table_one}
\end{table*}

Table \ref{tab:table_one}, which was generated using the Python software package tableone \cite{pollard_2018}, shows the baseline characteristics of the nonseptic and septic patient cohorts. For each patient, the maximum value of a feature was used. The \textit{Missing} column shows the number of total missing patient values in the dataset for a specific feature. There was no significant statistical difference between the average age for the nonsepsis and sepsis populations, which were 61.6 years and 62.1 years respectively.
\par In both the sepsis and nonsepsis populations, a majority of the patients were male. However, this disparity was larger in the sepsis population, with 59.9 percent of these patients being male, compared to the 55.6 percent of male patients in the nonsepsis population.
\par Another observation of the cohort was the large disparity in the hospital admission time between sepsis and nonsepsis patients. For sepsis patients, the median time between hospital admission and ICU admission (3.3 hours) was much lower when compared to nonsepsis patients (6 hours). This makes sense as patients that were at a greater risk of developing sepsis would need to be admitted to the ICU much quicker than patients at a lower risk for developing sepsis.
\par Another difference between the sepsis and nonsepsis cohorts was observed in the ICU length of stay. For sepsis patients, the average ICU length of stay was 67 hours, while for nonsepsis patients, the average ICU length of stay was 42 hours. This makes sense as patients that developed sepsis spent much more time in the ICU.
\par More significant differences between the sepsis and nonsepsis cohorts can be observed through the laboratory values. Generally, sepsis patients tended to have higher laboratory values than nonsepsis patients. This is to be expected since sepsis patients are in a poorer health state than nonsepsis patients. Additionally, the sepsis patients had more labs ordered over 12 hours and 48 hours than nonsepsis patients. This also seems reasonable as the sepsis patients would require more frequent monitoring and therefore would have labs ordered on a more frequent basis.

\subsection{Model Results}
\begin{table*}[h]
    \centering
    \begin{tabular}{p{45pt} p{40pt} p{40pt} p{40pt} p{40pt} p{40pt} p{40pt}} 
     \hline
     \textbf{Model} & \textbf{Specificity} & \textbf{Sensitivity} & \textbf{Precision} & \textbf{Accuracy} & \textbf{AUROC} & \textbf{AUPRC}  \\
     \hline
     RNN\_20lc & 0.6036 & 0.4857 & 0.0178 & 0.6021 & 0.5703 & 0.0188 \\

     RNN\_40lc & \textbf{0.7272} & 0.5711 & 0.0268 & \textbf{0.7252} & 0.7112 & 0.0316 \\

     RNN\_60lc & 0.7158 & 0.7152 & 0.0319 & 0.7158 & 0.7787 & 0.0442 \\

     RNN\_80lc & 0.7107 & 0.7659 & 0.0335 & 0.7114 & 0.7996 & 0.0484 \\

     RNN\_100lc & 0.717 & 0.7774 & 0.0347 & 0.7177 & 0.8097 & 0.0501 \\
     \hline
     RNN\_20m & 0.6117 & 0.4856 & 0.0182 & 0.6101 & 0.5735 & 0.019 \\

     RNN\_40m & 0.7209 & 0.5823 & 0.0267 & 0.7191 & 0.7123 & 0.0318 \\

     RNN\_60m & 0.7109 & 0.7209 & 0.0316 & 0.711 & 0.7784 & 0.0441 \\

     RNN\_80m & 0.7088 & 0.7699 & 0.0334 & 0.7096 & 0.7998 & 0.0482 \\

     RNN\_100m & 0.7114 & 0.7862 & 0.0345 & 0.7124 & 0.8101 & 0.05 \\
     \hline
     RNN\_20e & 0.6129 & 0.4809 & 0.0181 & 0.6112 & 0.5719 & 0.0189 \\

     RNN\_40e & 0.7258 & 0.578 & 0.027 & 0.7239 & 0.7135 & 0.0318 \\

     RNN\_60e & 0.7125 & 0.7182 & 0.0317 & 0.7126 & 0.7785 & 0.0439 \\
   
     RNN\_80e & 0.7116 & 0.767 & 0.0336 & 0.7123 & 0.8008 & 0.0485 \\

     RNN\_100e & 0.7143 & 0.7842 & 0.0347 & 0.7152& 0.8103 & \textbf{0.0503} \\
     \hline
     RNN	& 0.7105 & \textbf{0.7979} & \textbf{0.0348} & 0.7116 & \textbf{0.8192} & 0.05 \\
     \hline
    \end{tabular}
    \caption{Experiment results for the ALRt models and the baseline RNN model. Values represent the average of the 5-fold cross-validation.}
    \label{tab:results}
\end{table*}

The results on the test dataset are shown in Table \ref{tab:results}. From the table, it can be seen that the ALRt models with at least 60 percent of labeled training data achieve comparable performance on all metrics than the RNN model.
\par Regarding sensitivity, it can be seen that the ALRt models trained on more of the training data have higher sensitivity than the ALRt models trained on less of the training data. This is to be expected as more uncertain samples are added to the training set after each round of active learning.
\par Similar to sensitivity, the AUC and AUPRC values of the ALRt models trained on more of the training data were higher than those of the ALRt models trained on less of the training data. Another interesting observation is that the ALRt models trained on all of the data achieve performance that is most similar to that of the base RNN model. This makes sense since all these models utilize the entire dataset.

\begin{figure*}[h]
     \centering
     \begin{subfigure}[b]{0.3\textwidth}
         \centering
         \includegraphics[width=\textwidth]{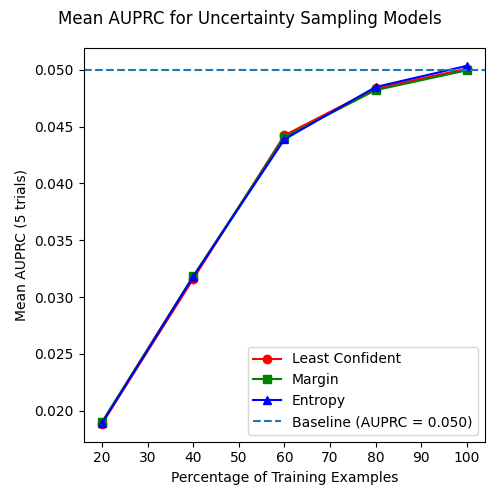}
     \end{subfigure}
     \begin{subfigure}[b]{0.3\textwidth}
         \centering
         \includegraphics[width=\textwidth]{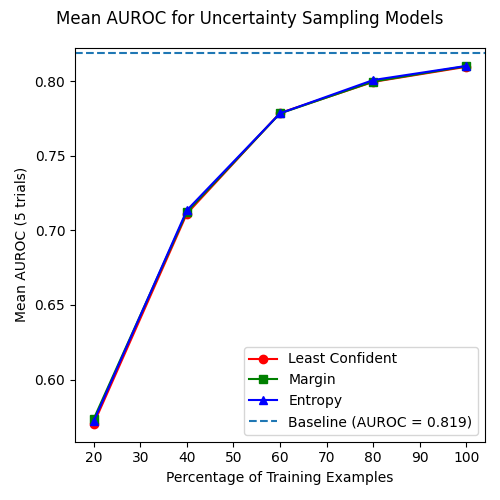}
     \end{subfigure}
     \begin{subfigure}[b]{0.3\textwidth}
         \centering
         \includegraphics[width=\textwidth]{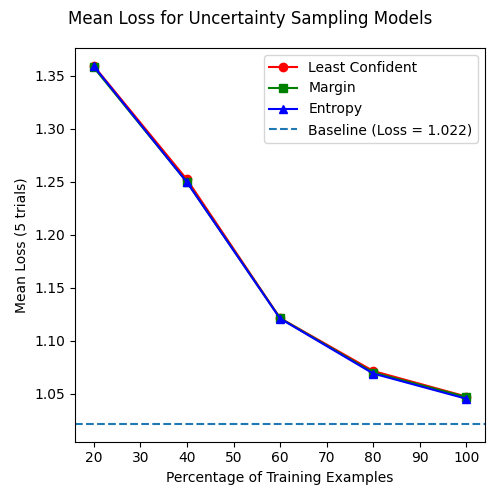}
     \end{subfigure}
     \caption{The AUPRC (left), AUROC (middle), and test loss (right) as a function of the amount of labeled training data}
     \label{fig:auprc_auroc_loss}
\end{figure*}

Figure \ref{fig:auprc_auroc_loss} shows the comparison of different metrics as a function of the amount of labeled training data. It can be seen that the AUPRC, AUROC, and loss all improve as the number of labeled training examples increases. Additionally, the AUPRC and AUROC reach about 90 percent of the baseline with only 60 percent of labeled training examples.

\subsection{Feature Importance Results}
Feature importance for the different RNN models was determined by using the Shapley Additive Explanations (SHAP) tool developed by \cite{lundberg2017}. The premise behind SHAP explanations is to determine the amount of positive or negative contribution of a given data point's input features to the model's output prediction. 
\par In the SHAP summary plot, the features are ordered in terms of feature importance according to the model, with the most important feature among all input data appearing at the top. The SHAP value of a given observation correlates to how that observation's feature value positively or negatively impacted the model's output prediction. The SHAP plots in Figures \ref{fig:shap_data_amount}, \ref{fig:shap_uncertainty}, and \ref{fig:shap_active_vs_base} display the 10 most important features for sepsis prediction according to each model.

\subsubsection{Amount of Labeled Training Data}
\begin{figure*}[h]
     \centering
     \begin{subfigure}[b]{0.3\textwidth}
         \centering
         \includegraphics[width=\textwidth]{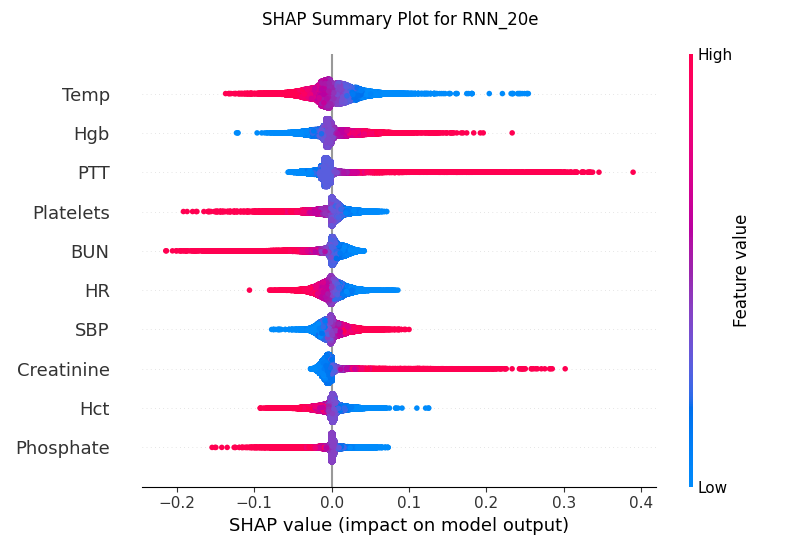}
     \end{subfigure}
     \begin{subfigure}[b]{0.3\textwidth}
         \centering
         \includegraphics[width=\textwidth]{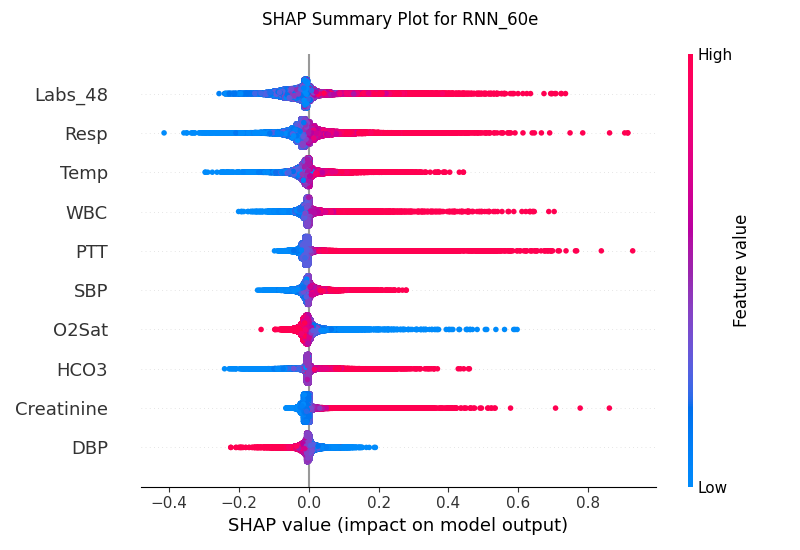}
     \end{subfigure}
     \begin{subfigure}[b]{0.3\textwidth}
         \centering
         \includegraphics[width=\textwidth]{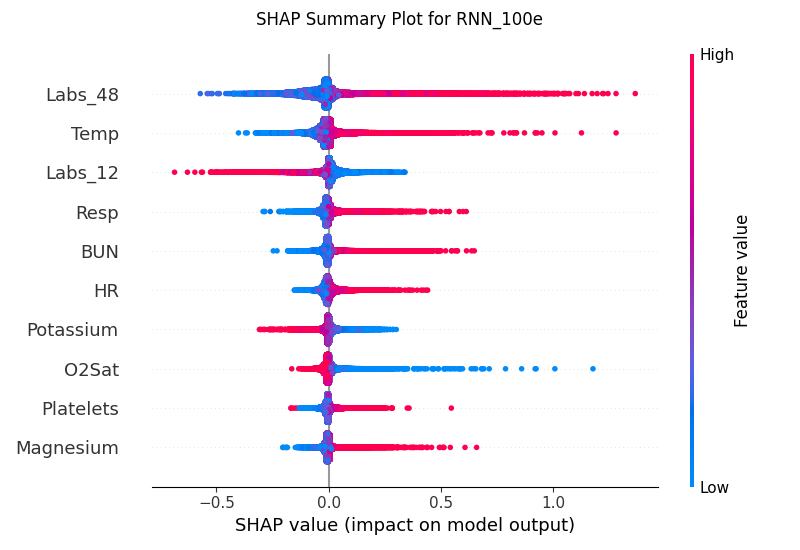}
     \end{subfigure}
     \caption{SHAP Plots for ALRt models trained on 20 percent (left), 60 percent (middle), 100 percent (right) of the training data}
     \label{fig:shap_data_amount}
\end{figure*}

Figure \ref{fig:shap_data_amount} shows the comparison of feature importance values across models using different amounts of labeled training data. Here, models using the entropy sampling method trained with different amounts of data are compared. Additionally, it can be seen that not only does the feature importance ranking varies between each model, but some features do not appear on every plot. This is most likely due to the fact that as training progresses, the model receives access to more training data. As a result, the model changes its feature importance rankings based on the newly available training data.

\subsubsection{Active Learning Sampling Criteria}
\begin{figure*}[h]
     \centering
     \begin{subfigure}[b]{0.3\textwidth}
         \centering
         \includegraphics[width=\textwidth]{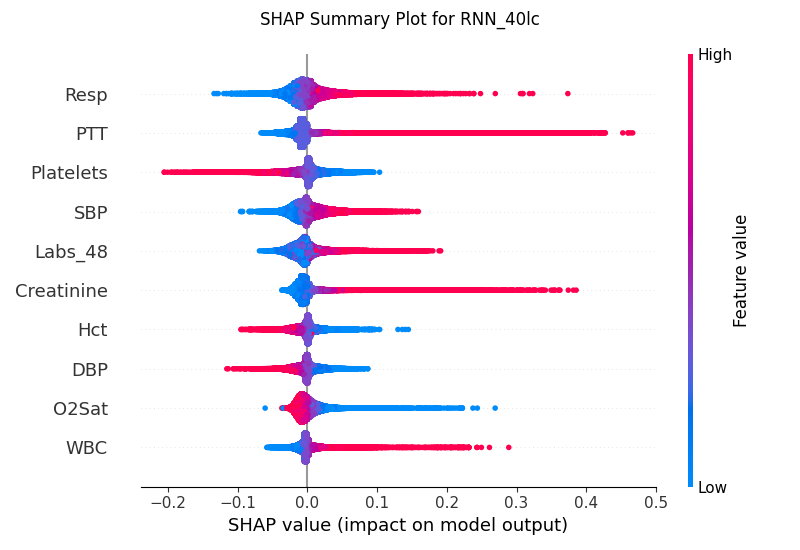}
     \end{subfigure}
     \begin{subfigure}[b]{0.3\textwidth}
         \centering
         \includegraphics[width=\textwidth]{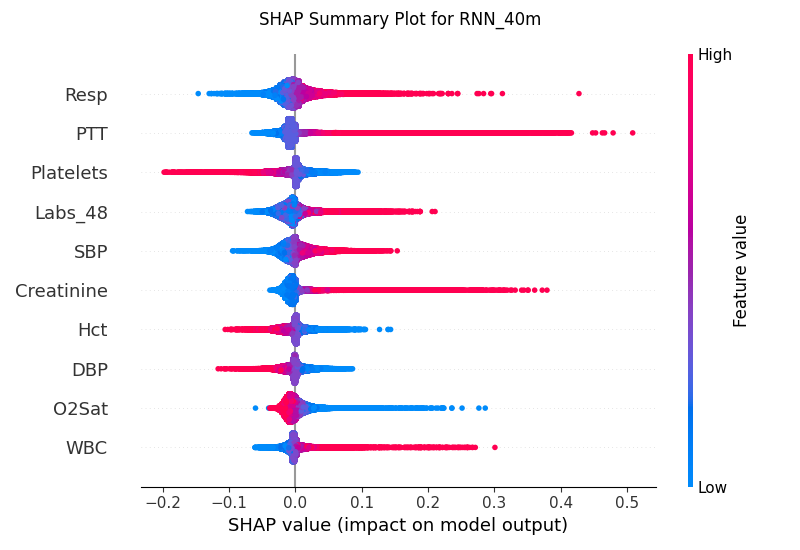}
     \end{subfigure}
     \begin{subfigure}[b]{0.3\textwidth}
         \centering
         \includegraphics[width=\textwidth]{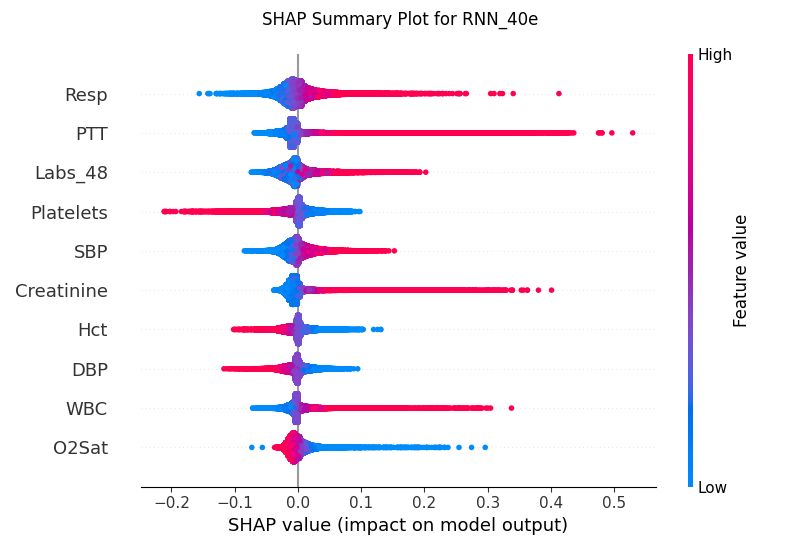}
     \end{subfigure}
     \caption{SHAP Plots for least confident (left), margin (middle), and entropy (right) sampling ALRt models}
     \label{fig:shap_uncertainty}
\end{figure*}

Figure \ref{fig:shap_uncertainty} shows the comparison of feature importance values across active learning models using different sampling criteria. These particular active learning models used 40 percent of the training data. It can be seen that although these models have many of the same features included in their SHAP plots, these features have different levels of importance rankings for each model. This is probably due to the difference in the samples selected by each uncertainty metric. As a result, these models will be trained on different data and therefore develop different feature importance rankings.

\subsubsection{Active Learning Framework vs Baseline Model}
\begin{figure*}[h]
     \centering
     \begin{subfigure}[b]{0.45\textwidth}
         \centering
         \includegraphics[width=\textwidth]{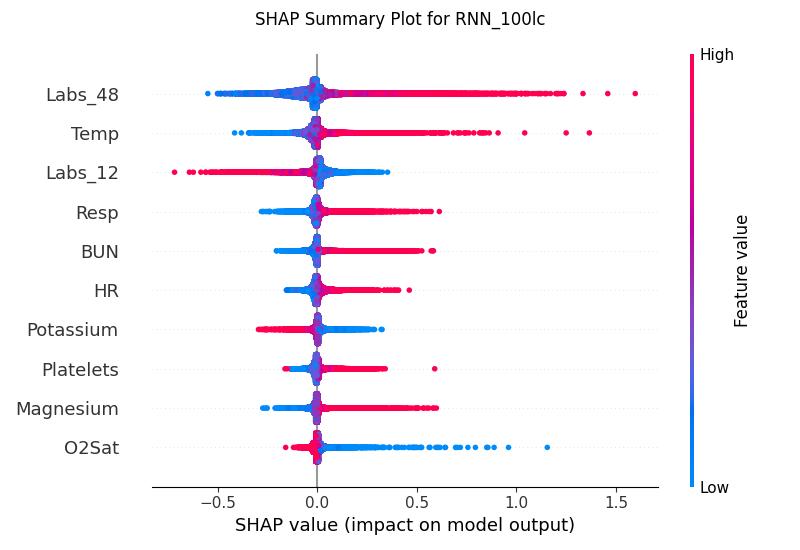}
     \end{subfigure}
     \begin{subfigure}[b]{0.45\textwidth}
         \centering
         \includegraphics[width=\textwidth]{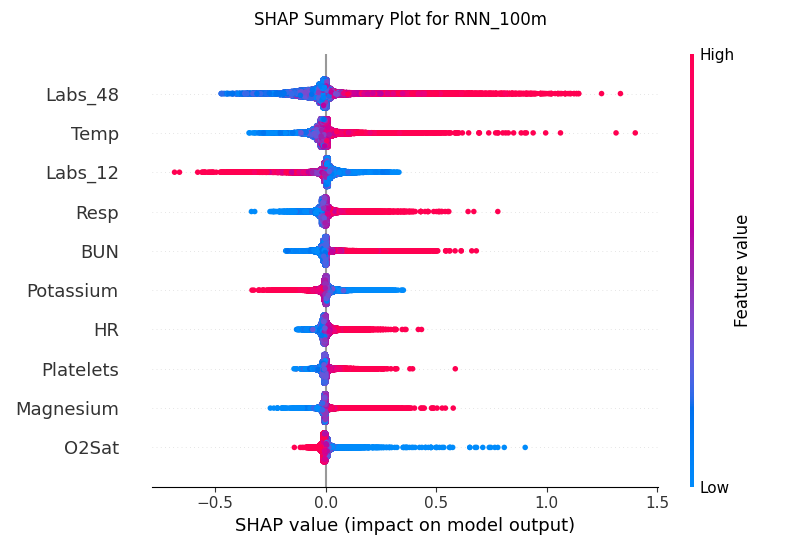}
     \end{subfigure}
     \hfill
     \begin{subfigure}[b]{0.45\textwidth}
         \centering
         \includegraphics[width=\textwidth]{shap_plots/shap_RNN_100e.png}
     \end{subfigure}
     \begin{subfigure}[b]{0.45\textwidth}
         \centering
         \includegraphics[width=\textwidth]{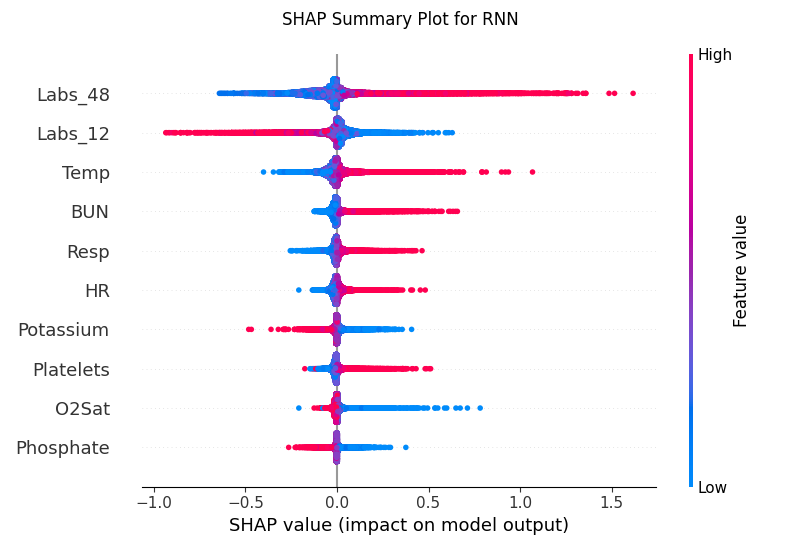}
     \end{subfigure}
     \caption{SHAP Plots of ALRts and Regular RNN}
     \label{fig:shap_active_vs_base}
\end{figure*}

Figure \ref{fig:shap_active_vs_base} shows the comparison of feature importance values in the fully trained ALRt models and the regularly trained model. A striking observation is that the fully trained uncertainty sampling and margin sampling ALRt models found many of the same features to be important in a similar order of importance. This makes sense, as the two active learning sampling techniques are fairly similar.

\section{Discussion}
\label{sec:discussion}
In the experiments, we sought to investigate whether active learning techniques could be used to develop accurate temporal prediction frameworks on limited labeled training data. Results indicate that the active learning framework provided comparable inference performance to a traditional temporal model using substantially more labeled training data.  

\subsection{Implications}
Unlike other sepsis prediction models \cite{nemati_interpretable_2018, desautels_prediction_2016, brown_prospective_2016, goh_artificial_2021}, this approach leverages the temporal aspects that are prevalent in electronic health records (EHR). Although other models have also utilized these temporal aspects \cite{futoma_learning_2017, futoma_improved_2017, scherpf_predicting_2019, zhang2021, moor2019, kok_automated_2020, wang_sepsis_2022}, this approach requires only limited training data to make robust predictions. The results confirm that the active learning framework can achieve comparable results to other prediction frameworks with fewer data. Active frameworks can assist in situations where adjudication resources are limited or expensive. Additionally, active frameworks may prove useful for rare disease classification tasks, where the number of labeled training data is often limited and the classes are highly imbalanced.

\subsection{Limitations}
One limitation of this work was that the data used for model training and evaluation was obtained from nondiverse hospital sites. Furthermore, many patients were older (64 years) and mostly male (55.8 percent). In the future, it would be worthwhile to utilize active learning methods with data from different hospitals with different patient distributions in order to evaluate the generalizability of these methods. For instance, it would be interesting to evaluate active learning methods on data from hospitals with patients that are younger and mostly female.
\par Another limitation of this work is that it focuses primarily on the viability of using active learning methods with recurrent neural networks for predictions on short temporal horizons. Future experiments would investigate the use of different active learning stopping criteria to develop a robust prediction model. This has been a popular research problem within active learning \cite{vlachos_stopping_2008, zhu2010}.

\section{Conclusion}
\label{sec:conclusion}
Sepsis is a deadly condition affecting many people throughout the world, with recent studies showing that the number of people diagnosed is steadily increasing. In this paper, we examined the use of active learning methods with RNNs in sepsis prediction. Our results show that active learning techniques for short temporal horizons can be used to develop robust prediction models for temporal events using limited training data. In the future, we hope to incorporate novel active learning methods to develop prediction models that generalize to various patient population distributions.

\bibliographystyle{unsrtnat}

\end{document}